\documentclass[runningheads]{llncs}
\usepackage{graphicx}
\usepackage{subfigure} 
\usepackage{epsfig}
\usepackage{amsmath}
\usepackage{amssymb} 
\usepackage{color}
\usepackage{url}
\usepackage{multirow}
\usepackage{hyperref}
\usepackage[width=122mm,left=12mm,paperwidth=146mm,height=193mm,top=12mm,paperheight=217mm]{geometry}
\begin{document}
\pagestyle{headings}
\mainmatter

\title{Attribute2Image: Conditional Image Generation from Visual Attributes} 

\titlerunning{Attribute2Image: Conditional Image Generation from Visual Attributes}

\authorrunning{Xinchen Yan, Jimei Yang, Kihyuk Sohn and Honglak Lee}

\author{Xinchen Yan$^1$, Jimei Yang$^2$, Kihyuk Sohn$^3$ and Honglak Lee$^1$}


\institute{$^1$Computer Science and Engineering,\\ 
	University of Michigan\\
	$^2$Adobe Research, $^3$NEC Labs\\
	\email{xcyan@umich.edu, jimyang@adobe.com, ksohn@nec-labs.com, honglak@umich.edu}
}

\maketitle
\begin{abstract}

This paper investigates a novel problem of generating images from visual attributes.
We model the image as a composite of foreground and background and develop a layered generative model with disentangled latent variables that can be learned end-to-end using a variational auto-encoder.
We experiment with natural images of faces and birds and demonstrate that the proposed models are capable of generating realistic and diverse samples with disentangled latent representations.
We use a general energy minimization algorithm for posterior inference of latent variables given novel images. 
Therefore, the learned generative models show excellent quantitative and visual results in the tasks of attribute-conditioned image reconstruction and completion.
%
\end{abstract}

\section{Introduction}
Generative image modeling is of fundamental interest in computer vision and machine learning.
Early works~\cite{smolensky1986information,srivastava2003advances,tu2007generative,lee2009convolutional,ranzato2010gatedmrf,le2011learning} studied statistical and physical principles of building generative models, but due to the lack of effective feature representations, their results are limited to textures or particular patterns such as well-aligned faces. 
Recent advances on representation learning using deep neural networks~\cite{krizhevsky2012imagenet,simonyan2014very} nourish a series of deep generative models that enjoy joint generative modeling and representation learning through Bayesian inference~\cite{TangS13,bengio2013deep,rezende2014stochastic,kingma2013auto,kingma2014semi,gregor2015draw} or adversarial training~\cite{goodfellow2014generative,denton2015deep}. 
Those works show promising results of generating natural images, but the generated samples are still in low resolution and far from being perfect because of the fundamental challenges of learning unconditioned generative models of images.

\begin{figure}[t]
\centering
\includegraphics[width=0.75\linewidth]{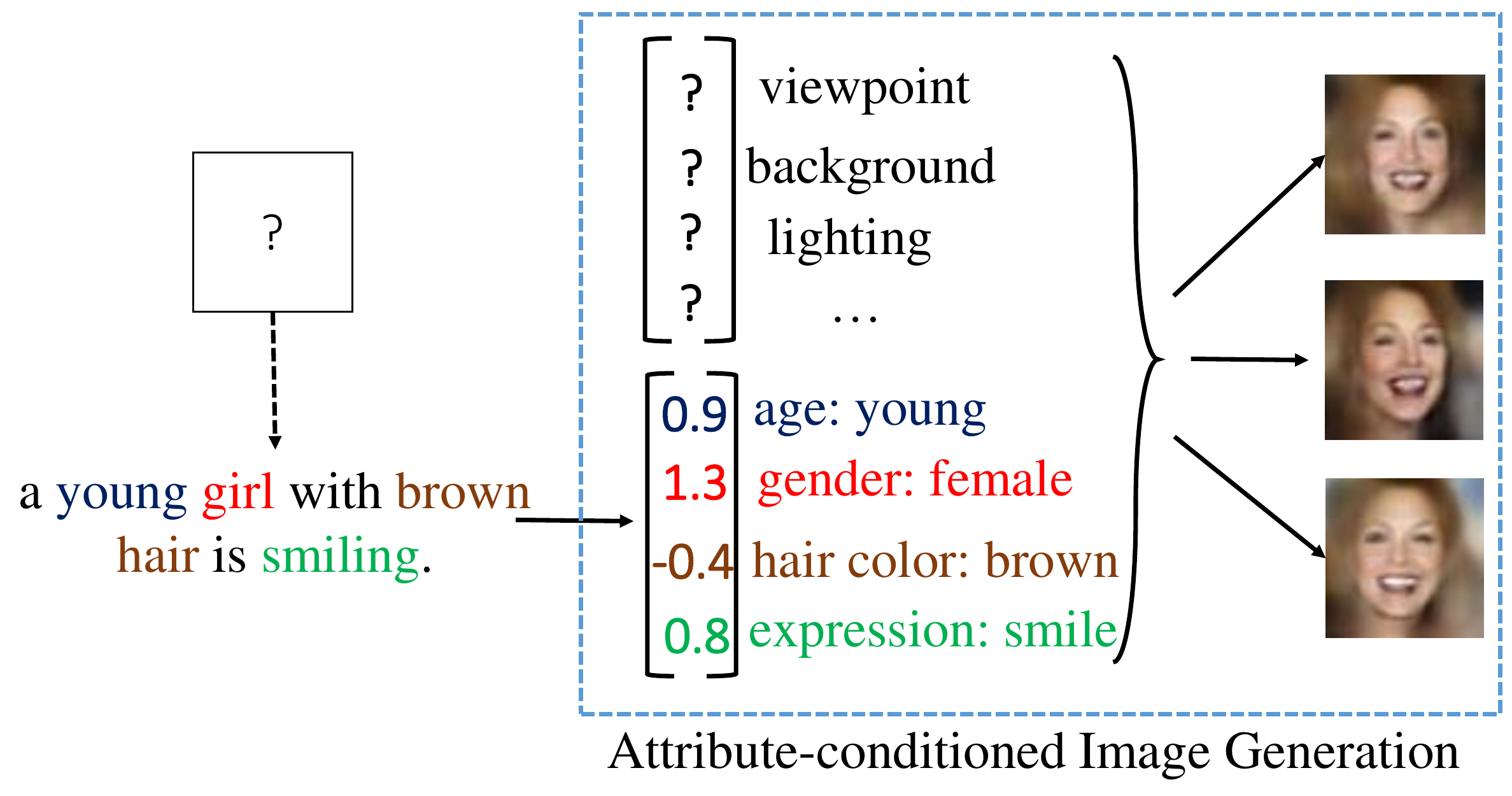}
\caption{An example that demonstrates the problem of conditioned image generation from visual attributes. We assume a vector of visual attributes is extracted from a natural language description, and then this attribute vector is combined with learned latent factors to generate diverse image samples.}
\label{figure:figure_intro}
\end{figure}

In this paper, we are interested in generating object images from high-level description. 
For example, we would like to generate portrait images that all match the description ``a young girl with brown hair is smiling'' (Figure~\ref{figure:figure_intro}).
This conditioned treatment reduces sampling uncertainties and helps generating more realistic images, and thus has potential real-world applications such as forensic art and semantic photo editing~\cite{Linder2013pixeltone,yang2011expresisonflow,kemelmacher-shlizerman2014age}.
The high-level descriptions are usually natural languages, but what underlies its corresponding images are essentially a group of facts or visual attributes that are extracted from the sentence.
In the example above, the attributes are (hair color: brown), (gender: female), (age: young) and (expression: smile).
Based on this assumption, we propose to learn an attribute-conditioned generative model.
Indeed, image generation is a complex process that involves many factors.
Other than enlisted attributes, there are many unknown or latent factors.
It has been shown that those latent factors are supposed to be interpretable according to their semantic or physical meanings~\cite{KulkarniWKT15,dosovitskiy2014learning,reed2014learning}. 
Inspired by layered image models~\cite{wang1994representing,nitzberg19992p1dsketch},
we disentangle the latent factors into two groups: one related to uncertain properties of foreground object and the other related to the background, and model the generation process as layered composition.
In particular, the foreground is overlaid on the background so that the background visibility depends on the foreground shape and position.
Therefore, we propose a novel layered image generative model with disentangled foreground and background latent variables. 
The entire background is first generated from background variables, then the foreground variables are combined with given attributes to generate object layer and its shape map determining the visibility of background and finally the image is composed by the summation of object layer and the background layer gated by its visibility map.
We learn this layered generative model in an end-to-end deep neural network using a variational auto-encoder~\cite{kingma2013auto} (Section~\ref{sec:method}).
Our variational auto-encoder includes two encoders or recognition models for approximating the posterior distributions of foreground and background latent variables respectively, and two decoders for generating a foreground image and a full image by composition.
Assuming the latent variables are Gaussian, the whole network can be trained end-to-end by back-propagation using the reparametrization trick.

Generating realistic samples is certainly an important goal of deep generative models. 
Moreover, generative models can be also used to perform Bayesian inference on novel images.
Since the true posterior distribution of latent variables is unknown, we propose a general optimization-based approach for posterior inference using image generation models and latent priors (Section~\ref{sec-posterior}).

We evaluate the proposed model on two datasets, the Labeled Faces in the Wild (LFW) dataset~\cite{huang2007labeled} and the Caltech-UCSD Birds-200-2011 (CUB) dataset~\cite{wah2011caltech}.
In the LFW dataset, the attributes are 73-dimensional vectors describing age, gender, expressions, hair and many others~\cite{kumar2009attribute}.
In the CUB dataset, the 312-dimensional binary attribute vectors are converted from descriptions about bird parts and colors. 
We organize our experiments in the following two tasks.
First, we demonstrate the quality of attribute-conditioned image generation with comparisons to nearest-neighbor search, and analyze the disentangling performance of latent space and corresponding foreground-background layers. 
Second, we perform image reconstruction and completion on a set of novel test images by posterior inference with quantitative evaluation.
Results from those experiments show the superior performance of the proposed model over previous art.
The contributions of this paper are summarized as follows:
\begin{itemize}
    \item We propose a novel problem of conditioned image generation from visual attributes.
    \item We tackle this problem by learning conditional variational auto-encoders and propose a novel layered foreground-background generative model that significantly improves the generation quality of complex images.
    \item We propose a general optimization-based method for posterior inference on novel images and use it to evaluate generative models in the context of image reconstruction and completion. 
\end{itemize}

\section{Related Work}
\paragraph{Image generation.}
In terms of generating realistic and novel images, there are several recent work~\cite{dosovitskiy2014learning,gregor2015draw,KulkarniWKT15,goodfellow2014generative,denton2015deep,radford2015unsupervised} that are relevant to ours.
Dosovitskiy et al.~\cite{dosovitskiy2014learning} proposed to generate 3D chairs given graphics code using deep convolutional neural networks, 
and Kulkarni et al.~\cite{KulkarniWKT15} used variational auto-encoders~\cite{kingma2013auto} to model the rendering process of 3D objects. 
Both of these models~\cite{KulkarniWKT15,dosovitskiy2014learning} assume the existence of a graphics engine during training, from which they have 1) virtually infinite amount of training data and/or 2) pairs of rendered images that differ only in one factor of variation. 
Therefore, they are not directly applicable to natural image generation.
While both work \cite{KulkarniWKT15,dosovitskiy2014learning} studied generation of rendered images from complete description (e.g., object identity, view-point, color) trained from synthetic images (via graphics engine), generation of images from an incomplete description (e.g., class labels, visual attributes) is still under-explored. 
In fact, image generation from incomplete description is a more challenging task and the one-to-one mapping formulation of \cite{dosovitskiy2014learning} is inherently limited.
Gregor et al.~\cite{gregor2015draw} developed recurrent variational auto-encoders with spatial attention mechanism that allows iterative image generation by patches. 
This elegant algorithm mimics the process of human drawing but at the same time faces challenges when scaling up to large complex images.
Recently, generative adversarial networks (GANs) ~\cite{goodfellow2014generative,gauthierconditional,denton2015deep,radford2015unsupervised} have been developed for image generation. 
In the GAN, two models are trained to against each other: a generative model aims to capture the data distribution, while a discriminative model attempts to distinguish between generated samples and training data. 
The GAN training is based on a min-max objective, which is known to be challenging to optimize.

\paragraph{Layered modeling of images.}
Layered models or 2.1D representations of images have been studied in the context of moving or still object segmentation \cite{wang1994representing,nitzberg19992p1dsketch,williams2004greedy,yang2012layered,isola2013scene}. 
The layered structure is introduced into generative image modeling~\cite{le2011learning,tang2012robust}.
Tang et al.~\cite{tang2012robust} modeled the occluded images with gated restricted Boltzmann machines and achieved good inpainting and denoising results on well cropped face images.
Le Roux et al.~\cite{le2011learning} explicitly modeled the occlusion layer in a masked restricted Boltzmann machine for separating foreground and background and demonstrated promising results on small patches.
Though similar to our proposed gating in the form, these models face  challenges when applied to model large natural images due to its difficulty in learning hierarchical representation based on restricted Boltzmann machine.


\paragraph{Multimodal Learning.}
%
%
%
%

Generative models of image and text have been studied in multimodal learning to model joint distribution of multiple data modalities~\cite{ngiam2011multimodal,srivastava2012multimodal,sohn2014improved}.
For example, Srivastava and Salakhutdinov~\cite{srivastava2012multimodal} developed a multimodal deep Boltzmann machine that models joint distribution of image and text (e.g., image tag). 
Sohn et al.~\cite{sohn2014improved} proposed improved shared representation learning of multimodal data through bi-directional conditional prediction by deriving a conditional prediction model of one data modality given the other and vice versa.
Both of these works focused more on shared representation learning using hand-crafted low-level image features and therefore have limited applications such as conditional image or text retrieval than actual generation of images.
%

\section{Attribute-conditioned Generative Modeling of Images}
\label{sec:method}
In this section, we describe our proposed method for attribute-conditioned generative modeling of images. 
%
We first describe a conditional variational auto-encoder, followed by the formulation of layered generative model and its variational learning.

\subsection{Base Model: Conditional Variational Auto-Encoder (CVAE)}
Given the attribute $y\in\mathbb{R}^{N_y}$ and latent variable $z\in\mathbb{R}^{N_z}$, our goal is to build a model $p_\theta (x | y,z)$ that generates realistic image $x\in\mathbb{R}^{N_x}$ conditioned on $y$ and $z$.
Here, we refer $p_\theta$ a generator (or generation model), parametrized by $\theta$. 
Conditioned image generation is simply a two-step process in the following:
\begin{enumerate}
    \item Randomly sample latent variable $z$ from prior distribution $p (z)$;
    \item Given $y$ and $z$ as conditioning variable, generate image $x$ from $p_\theta(x|y, z)$.
\end{enumerate}

Here, the purpose of learning is to find the best parameter $\theta$ that maximizes the log-likelihood $\log p_\theta (x|y)$.
%
As proposed in~\cite{rezende2014stochastic,kingma2013auto}, variational auto-encoders try to maximize the variational lower bound of the log-likelihood $\log p_{\theta}(x|y)$.
Specifically, an auxiliary distribution $q_{\phi}(z|x,y)$ is introduced to approximate the true posterior $p_{\theta}(z|x,y)$.
We refer the base model a conditional variational auto-encoder (CVAE) with the conditional log-likelihood
\begin{align}
\log p_\theta (x|y)
= \textit{KL} ( q_\phi (z|x,y) || p_\theta (z|x,y) ) + \mathcal{L}_{\text{CVAE}} (x,y;\theta, \phi),\nonumber
\end{align}
where the variational lower bound
\begin{align}
\mathcal{L}_{\text{CVAE}} (x,y;\theta, \phi) = -\textit{KL}(q_\phi (z|x,y) || p_\theta (z)) + \mathbb{E}_{q_\phi (z|x,y)} \big[ \log p_\theta (x|y,z) \big]\label{eqn:method-lb2}
\end{align}
is maximized for learning the model parameters.

Here, the prior $p_{\theta} (z) $ is assumed to follow isotropic multivariate Gaussian distribution, while two conditional distributions $p_\theta (x|y,z)$ and $q_\phi(z|x,y)$ are multivariate Gaussian distributions whose mean and covariance are parametrized by $\mathcal{N}\left(\mu_{\theta}(z,y), diag(\sigma^2_{\theta}(z,y))\right)$ and $\mathcal{N}\left(\mu_\phi (x, y), diag(\sigma^2_\phi(x, y))\right)$, respectively. 
%
We refer the auxiliary proposal distribution $q_\phi(z|x,y)$ a recognition model and the conditional data distribution $p_{\theta} (x|y,z)$ a generation model.

The first term $\textit{KL}(q_\phi(z|x,y)||p_\theta(z))$ is a regularization term that reduces the gap between the prior $p(z)$ and the proposal distribution $q_\phi(z|x,y)$, while the second term $\log p_\theta (x|y,z)$ is the log likelihood of samples.
In practice, we usually take as a deterministic generation function the mean $x=\mu_{\theta}(z,y)$ of conditional distribution $p_{\theta}(x|z,y)$ given $z$ and $y$, so it is convenient to assume the standard deviation function $\sigma_{\theta}(z,y)$ is a constant shared by all the pixels as the latent factors capture all the data variations.
We will keep this assumption for the rest of the paper if not particularly mentioned.
%
Thus, we can rewrite the second term in the variational lower bound as reconstruction loss $L(\cdot, \cdot)$ (e.g., $\ell_2$ loss):
\begin{align}
\mathcal{L}_{\text{CVAE}} = & -\textit{KL} ( q_\phi (z|x,y) || p_\theta (z) ) - \mathbb{E}_{q_\phi (z|x,y)} L(\mu_\theta(y,z), x)
\end{align}
Note that the discriminator of GANs~\cite{goodfellow2014generative} can be used as the loss function $L(\cdot, \cdot)$ as well, especially when $\ell_2$ (or $\ell_1$) reconstruction loss may not capture the true image similarities.
We leave it for future study.

\subsection{Disentangling CVAE with a Layered Representation}
\begin{figure*}[t]
\centering
\subfigure[CVAE: $p_{\theta}(x|y,z)$]{\includegraphics[height=1.0in]{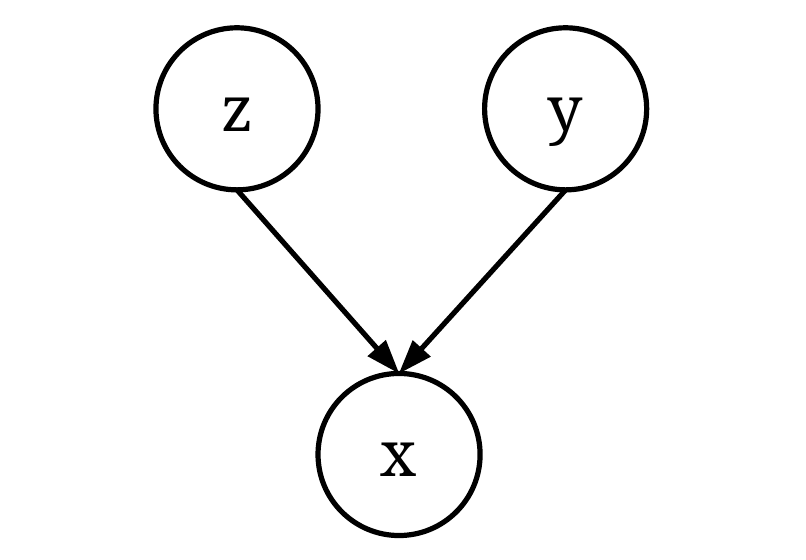}}\hspace{0.1in}
\subfigure[disCVAE: $p_{\theta}(x, x_{F}, g|y, z_{F}, z_{B})$]{\includegraphics[height=1.0in]{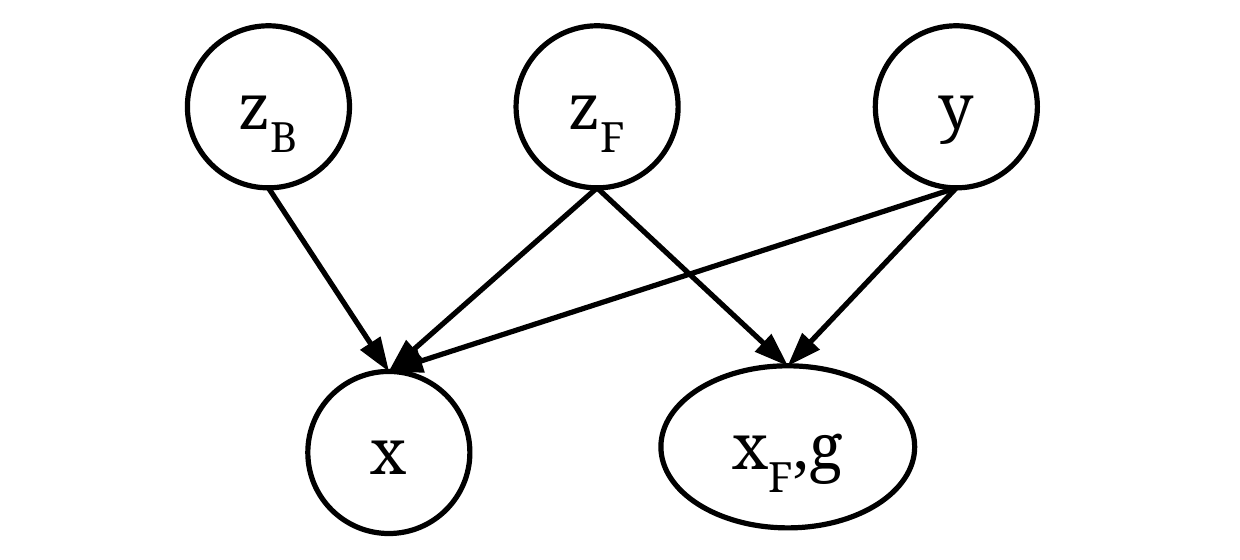}}
\caption{ Graphical model representations of attribute-conditioned image generation models (a) without (CVAE) and (b) with (disCVAE) disentangled latent space.}
\label{figure:figure_gm}
\end{figure*}
An image $x$ can be interpreted as a composite of a foreground layer (or a foreground image $x_F$) and a background layer (or a background image $x_B$) via a matting equation~\cite{porter1984compositing}:
\begin{align}
x = x_F \odot (1 - g) + x_B \odot g,
\label{eqn:compose1}
\end{align}
where $\odot$ denotes the element-wise product.
$g \in [0,1]^{N_x}$ is an occlusion layer or a gating function that determines the visibility of background pixels while $1-g$ defines the visibility of foreground pixels. 
%
%
However, the model based on Equation~\eqref{eqn:compose1} may suffer from the incorrectly estimated mask as it gates the foreground region with imperfect mask estimation. 
Instead, we approximate the following formulation that is more robust to estimation error on mask:
\begin{align}
x = x_F + x_B \odot g.
\label{eqn:compose2}
\end{align}
When lighting condition is stable and background is at a distance, we can safely assume foreground and background pixels are generated from independent latent factors.
To this end, we propose a disentangled representation $z = [z_F, z_B]$ in the latent space, where $z_F$ together with attribute $y$ captures the foreground factors while $z_B$ the background factors.
As a result, the foreground layer $x_F$ is generated from $\mu_{\theta_F} (y,z_F)$ and the background layer $x_B$ from $\mu_{\theta_B} (z_B)$.
The foreground shape and position determine the background occlusion so the gating layer $g$ is generated from $s_{\theta_g} (y,z_F)$ where the last layer of $s(\cdot)$ is sigmoid function.
%
In summary, we approximate the layered generation process as follows: 
\begin{enumerate}
    \item Sample foreground and background latent variables $z_F\sim p (z_F)$, $z_B\sim p (z_B)$;
\item Given $y$ and $z_{F}$, generate foreground layer $x_{F}\sim\mathcal{N}\left(\mu_{\theta_{F}}(y,z_{F}),\sigma_{0}^{2} I_{N_{x}} \right)$
and gating layer $g\sim Bernoulli\left(s_{\theta_{g}}(y,z_{F})\right)$; 
here, $\sigma_{0}$ is a constant. The background layer (which correspond
to $x_{B}$) is implicitly computed as $\mu_{\theta_{B}}(z_{B})$.
\item Synthesize an image $x\sim\mathcal{N}\left(\mu_{\theta}(y,z_{F},z_{B}),\sigma_{0}^{2} I_{N_{x}} \right)$
where $\mu_{\theta}(y,z_{F},z_{B})=\mu_{\theta_{F}}(y,z_{F})+s_{\theta_{g}}(y,z_{F})\odot\mu_{\theta_{B}}(z_{B})$.
\end{enumerate}
\paragraph{Learning.} 
%
It is very challenging to learn our layered generative model in a fully-unsupervised manner since we need to infer about $x_F$, $x_B$, and $g$ from the image $x$ only.
In this paper, we further assume the foreground layer $x_F$ (as well as gating variable $g$) is observable during the training and we train the model to maximize the joint log-likelihood $\log p_{\theta} (x,x_F,g|y)$ instead of $\log p_\theta (x|y)$. 
With disentangled latent variables $z_F$ and $z_B$, we refer our layered model a disentangling conditional variational auto-encoder (disCVAE). 
We compare the graphical models of disCVAE with vanilla CVAE in Figure~\ref{figure:figure_gm}.
Based on the layered generation process, we write the generation model by
\begin{align}
p_\theta (x_F, g, x, z_F, z_B|y) =&\; p_\theta (x|z_F, z_B, y) p_\theta (x_F,g|z_F, y) p_\theta (z_F) p_\theta (z_B),
\end{align}
the recognition model by 
\begin{align}
q_\phi (z_F, z_B|x_F,g,x,y) =&\; q_\phi (z_B|z_F, x_F, g, x, y) q_\phi (z_F|x_F, g, y)
\end{align}
and the variational lower bound $\mathcal{L}_{\text{disCVAE}} (x_F,g,x,y;\theta, \phi)$ is given by
\begin{align}
&\; \mathcal{L}_{\text{disCVAE}} (x_F,g,x,y;\theta, \phi)=\nonumber\\
&\; -\textit{KL} ( q_\phi (z_F|x_F,g,y) || p_\theta(z_F) ) 
- \mathbb{E}_{q_\phi (z_F|x_F,g,  y)} \big[ \textit{KL} ( q_\phi (z_B|z_F, x_F, g, x, y) || p_\theta (z_B) ) \big]\nonumber\\
&\ - \mathbb{E}_{q_\phi (z_F|x_F, g, y)} \big[L(\mu_{\theta_F}(y, z_F), x_F) + \lambda_g L(s_{\theta_g}(y, z_F), g) \big]\nonumber\\
&\ - \mathbb{E}_{q_\phi(z_F, z_B|x_F, g, x, y)} L(\mu_{\theta} (y, z_F, z_B), x)
\end{align}
where $\mu_{\theta} (y, z_F, z_B)=\mu_{\theta_F}(y, z_F) + s_{\theta_g}(y, z_F)\odot\mu_{\theta_B}(z_B)$ as in Equation~\eqref{eqn:compose2}.
We further assume that $\log p_\theta(x_F, g|z_F, y) = \log p_\theta(x_F|z_F, y) + \lambda_g \log p_\theta(g|z_F, y)$, where we introduce $\lambda_g$ as additional hyperparameter when decomposing the probablity $p_\theta (x_F,g|z_F, y)$.
For the loss function $L(\cdot, \cdot)$, we used reconstruction error for predicting $x$ or $x_F$ and cross entropy for predicting the binary mask $g$. 
See the supplementary material~\ref{sec-app-deriv} for details of the derivation.
All the generation and recognition models are parameterized by convolutional neural networks and trained end-to-end in a single architecture with back-propagation. 
We will introduce the exact network architecture in the experiment section.

\section{Posterior Inference via Optimization}
\label{sec-posterior}
Once the attribute-conditioned generative model is trained, the inference or generation of image $x$ given attribute $y$ and latent variable $z$ is straight-forward. 
However, the inference of latent variable $z$ given an image $x$ and its corresponding attribute $y$ is unknown.
In fact, the latent variable inference is quite useful as it enables model evaluation on novel images.
For simplicity, we introduce our inference algorithm based on the vanilla CVAE and the same algorithm can be directly applied to the proposed disCVAE and the other generative models such as GANs~\cite{gauthierconditional,denton2015deep}. 
Firstly we notice that the recognition model $q_\phi (z|y,x)$ may not be directly used to infer $z$.
On one hand, as an approximate, we don't know how far it is from the true posterior $p_\theta (z|x,y)$ because the KL divergence between them is thrown away in the variational learning objective; on the other hand, this approximation does not even exist in the models such as GANs.
We propose a general approach for posterior inference via optimization in the latent space.
Using Bayes' rule, we can formulate the posterior inference by 
\begin{align}
\max_z \log p_\theta (z|x,y) =&\; \max_z \big[ \log p_\theta (x|z,y) + \log p_\theta (z|y) \big] \nonumber\\
=&\; \max_z \big[ \log p_\theta (x|z,y) + \log p_\theta (z) \big]
\label{eqn:inference}
\end{align}
Note that the generation models or likelihood terms $p_\theta (x|z,y)$ could be non-Gaussian or even a deterministic function (e.g. in GANs) with no proper probabilistic definition.
Thus, to make our algorithm general enough, we reformulate the inference in~\eqref{eqn:inference} as an energy minimization problem,
\begin{align}
\min_{z} E(z,x,y) = &\; \min_{z} \big[L(\mu(z,y), x) + \lambda R(z)\big]
\label{eqn:optim}
\end{align}
where $L(\cdot,\cdot)$ is the image reconstruction loss and $R(\cdot)$ is a prior regularization term.
Taking the simple Gaussian model as an example, the posterior inference can be re-written as,
\begin{align}
\min_{z} E(z,x,y) = &\; \min_{z} \big[\|\mu(z,y)-x\|^2 + \lambda \|z\|^2)\big]
\label{eqn:ell2}
\end{align}
Note that we abuse the mean function $\mu(z,y)$ as a general image generation function. 
Since $\mu(z,y)$ is a complex neural network, optimizing~\eqref{eqn:optim} is essentially error back-propagation from the energy function to the variable $z$, which we solve by the ADAM method~\cite{kingma2014adam}.
Our algorithm actually shares a similar spirit with recently proposed neural network visualization~\cite{yosinski2015understanding} and texture synthesis algorithms~\cite{gatys2015texture}. 
The difference is that we use generation models for recognition while their algorithms use recognition models for generation.
Compared to the conventional way of inferring $z$ from recognition model $q_\phi(z|x,y)$, the proposed optimization contributed to an empirically more accurate latent variable $z$ and hence was useful for reconstruction, completion, and editing.

\section{Experiments}
\label{sec-experiments}

\paragraph{Datasets.}
We evaluated our model on two datasets: Labeled Faces in the Wild (LFW)~\cite{huang2007labeled} and Caltech-UCSD Birds-200-2011 (CUB)~\cite{wah2011caltech}. 
For experiments on LFW, we aligned the face images using five landmarks~\cite{zhu2014transferring} and rescaled the center region to $64\times64$.
We used $73$ dimensional attribute score vector provided by~\cite{kumar2009attribute} that describes different aspects of facial appearance such as age, gender, or facial expression.
We trained our model using 70\% of the data (9,000 out of 13,000 face images) following the training-testing split (View 1)~\cite{huang2007labeled}, where the face identities are distinct between train and test sets.
For experiments on CUB, we cropped the bird region using the tight bounding box computed from the foreground mask and rescaled to $64\times64$.
We used $312$ dimensional binary attribute vector that describes bird parts and colors.
We trained our model using 50\% of the data (6,000 out of 12,000 bird images) following the training-testing split~\cite{wah2011caltech}.
For model training, we held-out 10\% of training data for validation.

\paragraph{Data preprocessing and augmentation.} To make the learning easier, we preprocessed the data by normalizing the pixel values to the range ${\left[-1, 1\right]}$.
We augmented the training data with the following image transformations~\cite{krizhevsky2012imagenet,eigen2014depth}:
1) flipping images horizontally with probability $0.5$, 2) multiplying pixel values of each color channel with a random value $c \in {\left[0.97, 1.03\right]}$, and 3) augmenting the image with its residual with a random tradeoff parameter $s \in {\left[0, 1.5\right]}$.
Specifically, for CUB experiments, we performed two extra transformations: 4) rotating images around the centering point by a random angle $\theta_r \in {\left[-0.08, 0.08\right]}$, 5) rescaling images to the scale of $72\times72$ and performing random cropping of $64\times64$ regions.
Note that these methods are designed to be invariant to the attribute description.

\paragraph{Architecture design.}
For disCVAE, we build four convolutional neural networks (one for foreground and the other for background for both recognition and generation networks) for auto-encoding style training.
%
%
%
The foreground encoder network consists of 5 convolution layers, followed by 2 fully-connected layers (convolution layers have 64, 128, 256, 256 and 1024 channels with filter size of $5 \times 5$, $5 \times 5$, $3 \times 3$, $3 \times 3$ and $4 \times 4$, respectively; the two fully-connected layers have 1024 and 192 neurons). The attribute stream is merged with image stream at the end of the recognition network.
The foreground decoder network consists of 2 fully-connected layers, followed by 5 convolution layers with 2-by-2 upsampling (fully-connected layers have 256 and $8 \times 8 \times 256$ neurons; the convolution layers have 256, 256, 128, 64 and 3 channels with filter size of $3 \times 3$, $5 \times 5$, $5 \times 5$, $5 \times 5$ and $5 \times 5$. The foreground prediction stream and gating prediction stream are separated at the last convolution layer.
We adopt the same encoder/decoder architecture for background networks but with fewer number of channels. 
See the supplementary material~\ref{sec-app-netarch} for more details.

For all the models, we fixed the latent dimension to be 256 and found this configuration is sufficient to generate $64 \times 64$ images in our setting.
We adopt slightly different architectures for different datasets: we use 192 dimensions to foreground latent space and 64 dimensions to background latent space for experiments on LFW dataset; we use 128 dimensions for both foreground and background latent spaces on CUB dataset. 
Compared to vanilla CVAE, the proposed disCVAE has more parameters because of the additional convolutions introduced by the two-stream architecture. However, we found that adding more parameters to vanilla CVAE does not lead to much improvement in terms of image quality. 
Although both \cite{dosovitskiy2014learning} and the proposed method use segmentation masks as supervision, naive mask prediction was not comparable to the proposed model in our setting based on the preliminary results.
In fact, the proposed disCVAE architecture assigns foreground/background generation to individual networks and composite with gated interaction, which we found very effective in practice. 

%
%

\paragraph{Implementation details.}
We used ADAM~\cite{kingma2014adam} for stochastic optimization in all experiments.
For training, we used mini-batch of size 32 and the learning rate $0.0003$.
We also added dropout layer of ratio $0.5$ for the image stream of the encoder network before merging with attribute stream.
For posterior inference, we used the learning rate $0.3$ with $1000$ iterations.
The models are implemented using deep learning toolbox Torch7~\cite{collobert2011torch7}.

\paragraph{Baselines.}
For the vanilla CVAE model, we used the same convolution architecture from foreground encoder network and foreground decoder network.
To demonstrate the significance of attribute-conditioned modeling, we trained an unconditional variational auto-encoders (VAE) with almost the same convolutional architecture as our CVAE.
%


\begin{figure*}[t]
\centering
\includegraphics[width=0.835\linewidth]{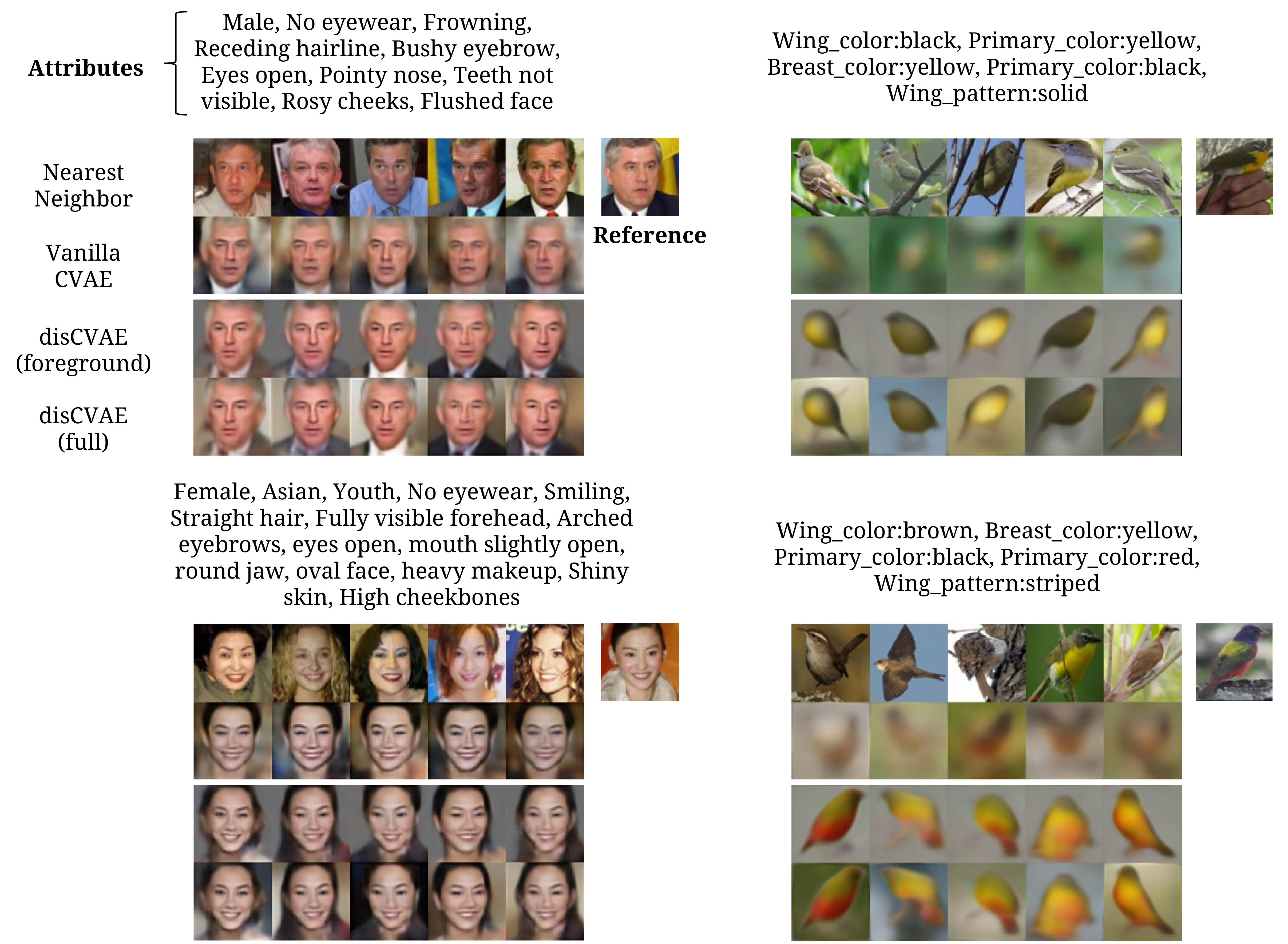}
\caption{Attribute-conditioned image generation.}
\label{figure:figure_condgen}
\end{figure*}
%
\subsection{Attribute-conditioned Image Generation}
To examine whether the model has the capacity to generate diverse and realistic images from given attribute description,
we performed the task of attribute-conditioned image generation.
For each attribute description from testing set, we generated 5 samples by the proposed generation process: $x \sim p_\theta (x|y, z)$, where $z$ is sampled from isotropic Gaussian distribution.
For vanilla CVAE, $x$ is the only output of the generation.
In comparison, for disCVAE, the foreground image $x_F$ can be considered a by-product of the layered generation process.
For evaluation, we visualized the samples generated from the model in Figure~\ref{figure:figure_condgen} and compared them with the corresponding image in the testing set, which we name as ``reference'' image.
To demonstrate that model did not exploit the trivial solution of attribute-conditioned generation by memorizing the training data, we added a simple baseline as experimental comparison. 
Basically, for each given attribute description in the testing set, we conducted the nearest neighbor search in the training set.
We used the mean squared error as the distance metric for the nearest neighbor search (in the attribute space).
For more visual results and code, please refer to the project website: \url{https://sites.google.com/site/attribute2image/}.

\paragraph{Attribute-conditioned face image generation.}
As we can see in Figure~\ref{figure:figure_condgen}, face images generated by the proposed models look realistic and non-trivially different from each other, especially for view-point and background color.
%
Moreover, it is clear that images generated by disCVAE have clear boundaries against the background.
In comparison, the boundary regions between the hair area and background are quite blurry for samples generated by vanilla CVAE. 
This observation suggests the limitation of vanilla CVAE in modeling hair pattern for face images.
This also justifies the significance of layered modeling and latent space disentangling in our attribute-conditioned generation process. 
Compared to the nearest neighbors in the training set, the generated samples can better reflect the input attribute description. For quantitative evaluations, please refer to supplementary material~\ref{sec-app-attrpred} for details.

\paragraph{Attribute-conditioned bird image generation.} 

Compared to the experiments on LFW database, the bird image modeling is more challenging because the bird images have more diverse shapes and color patterns and the binary-valued attributes are more sparse and higher dimensional.
As we can see in Figure~\ref{figure:figure_condgen}, there is a big difference between two versions of the proposed CVAE model.
Basically, the samples generated by vanilla CVAE are blurry and sometimes blended with the background area.
%
%
However, samples generated by disCVAE have clear bird shapes and reflect the input attribute description well.
This confirms the strengths of the proposed layered modeling of images.

\paragraph{Attribute-conditioned Image Progression.}

%
%
To better analyze the proposed model, we generate images with interpolated attributes by gradually increasing or decreasing the values along each attribute dimension. 
We regard this process as \textit{attribute-conditioned image progression}.
Specifically, for each attribute vector,
we modify the value of one attribute dimension by interpolating between the minimum and maximum attribute value.
Then, we generate images by interpolating the value of $y$ between the two attribute vectors while keeping latent variable $z$ fixed.
For visualization, we use the attribute vector from testing set.

\begin{figure}[t]
\centering
\includegraphics[width=0.95\linewidth]{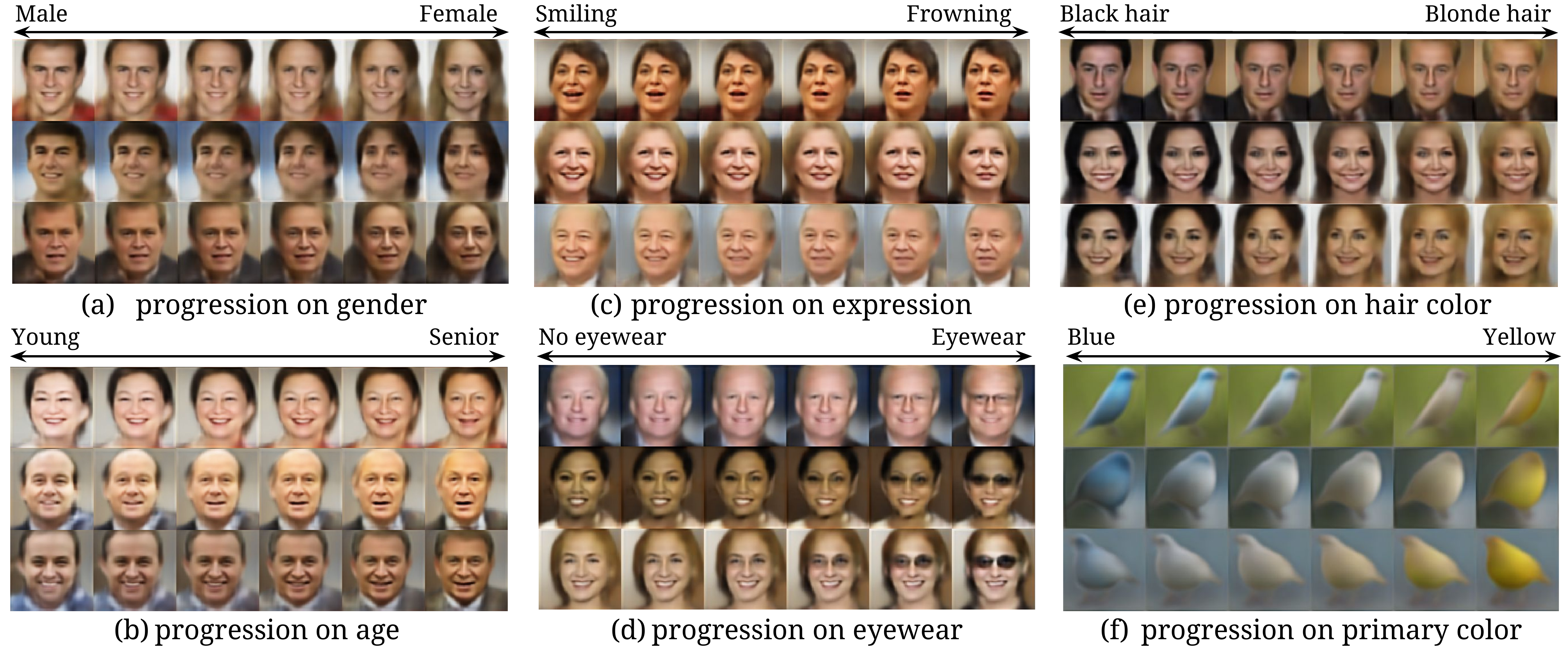}
\caption{Attribute-conditioned image progression.
The visualization is organized into six attribute groups (e.g., ``gender'', ``age'', ``facial expression'', ``eyewear'', ``hair color'' and ``primary color (blue vs. yellow)''). 
Within each group, the images are generated from $p_\theta(x|y,z)$ with $z \sim \mathcal{N}(0,I)$ and $y = [y_\alpha,y_{rest}]$, where $y_\alpha = (1-\alpha) \cdot y_{min} + \alpha \cdot y_{max}$. 
Here, $y_{min}$ and $y_{max}$ stands for the minimum and maximum attribute value respectively in the dataset along the corresponding dimension.
}
\label{figure:figure_prog_subset}
\end{figure}

As we can see in Figure~\ref{figure:figure_prog_subset}, samples generated by progression are visually consistent with attribute description. 
For face images, by changing attributes like ``gender'' and ``age'', the identity-related visual appearance is changed accordingly but the viewpoint, background color, and facial expression are well preserved; 
on the other hand, by changing attributes like ``facial expression'',``eyewear'' and ``hair color'', the global appearance is well preserved but the difference appears in the local region. 
For bird images, by changing the primary color from one to the other, the global shape and background color are well preserved.
These observations demonstrated that the generation process of our model is well controlled by the input attributes.

\begin{figure*}[t]
\centering
\includegraphics[width=.85\linewidth]{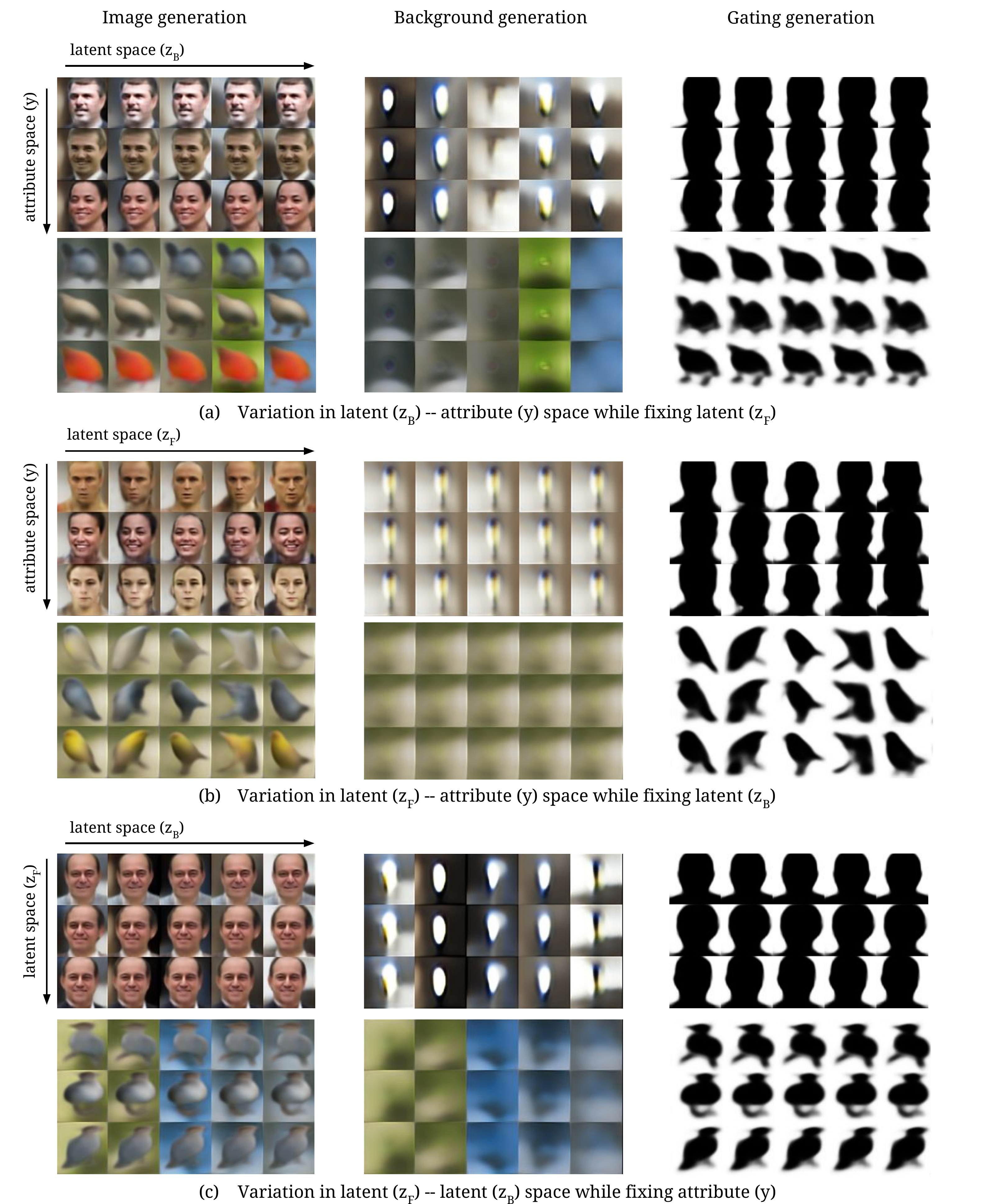}
\caption{Analysis: Latent Space Disentangling.}
\label{figure:figure_distengle}
\end{figure*}

\paragraph{Analysis: Latent Space Disentangling.}
To better analyze the disCVAE, we performed the following experiments on the latent space.
In this model, the image generation process is driven by three factors: attribute $y$, foreground latent variable $z_F$ and background latent variable $z_B$.
By changing one variable while fixing the other two, we can analyze how each variable contributes to the final generation results.
We visualize the samples $x$, the generated background $x_B$ and the gating variables $g$ in Figure~\ref{figure:figure_distengle}.
We summarized the observations as follows:
1) The background of the generated samples look different but with identical foreground region when we change background latent variable $z_B$ only; 2) the foreground region of the generated samples look diverse in terms of viewpoints but still look similar in terms of appearance and the samples have uniform background pattern when we change foreground latent variable $z_F$ only. 
Interestingly, for face images, one can identify a ``hole'' in the background generation. This can be considered as the location prior of the face images, since the images are relatively aligned.
Meanwhile, the generated background for birds are relatively uniform, which demonstrates our model learned to recover missing background in the training set and also suggests that foreground and background have been disentangled in the latent space.

\subsection{Attribute-conditioned Image Reconstruction and Completion}
%
%
\paragraph{Image reconstruction.} Given a test image $x$ and its attribute vector $y$, we find $z$ that maximizes the posterior $p_\theta (z|x, y)$ following Equation~\eqref{eqn:optim}. 
\paragraph{Image completion.} Given a test image with synthetic occlusion, we evaluate whether the model has the capacity to fill in the occluded region by recognizing the observed region.
We denote the occluded (unobserved) region and observed region as $x_u$ and $x_o$, respectively.
For completion, we first find $z$ that maximizes the posterior $p_\theta (z|x_o, y)$ by optimization~\eqref{eqn:optim}.
Then, we fill in the unobserved region $x_u$ by generation using $p_\theta (x_u|z,y)$.
For each face image, we consider four types of occlusions: occlusion on the eye region, occlusion on the mouth region, occlusion on the face region and occlusion on right half of the image. For occluded regions, we set the pixel value to 0.
%
For each bird image, we consider blocks of occlusion of size $8\times 8$ and $16 \times 16$ at random locations.

%

\begin{figure*}[b]
\centering
\includegraphics[width=1.0\linewidth]{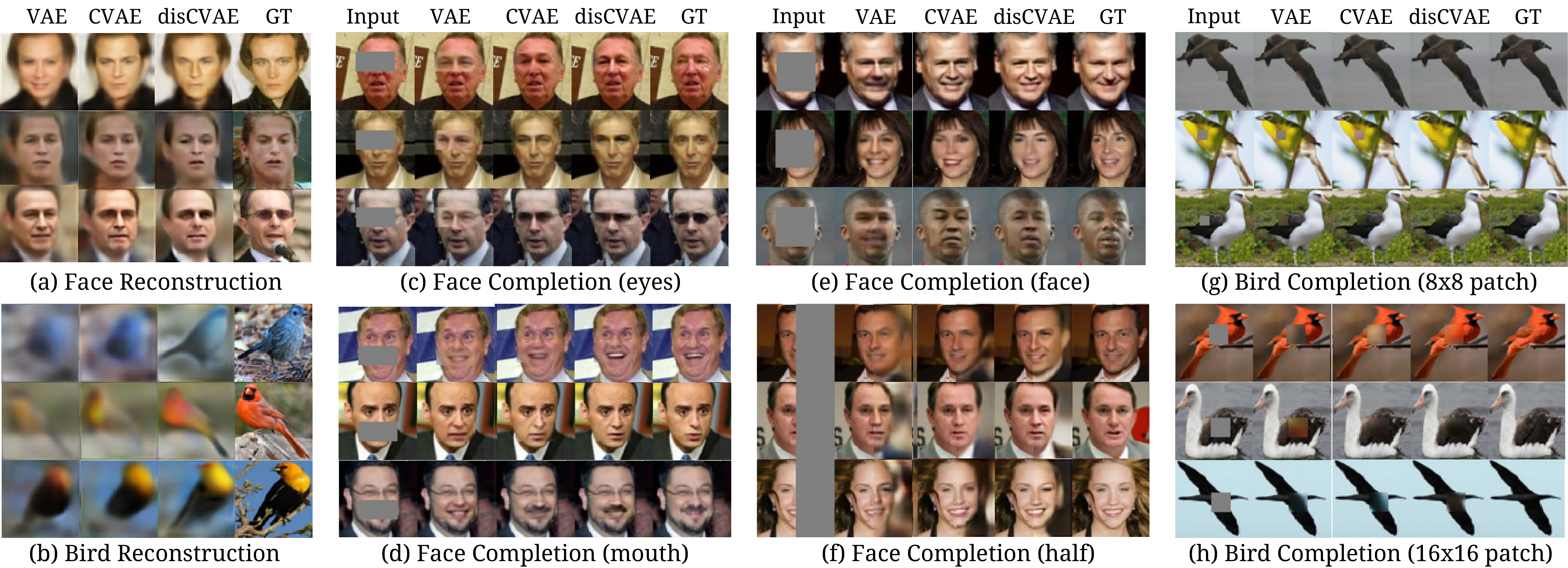}
\caption{ Attribute-conditioned image reconstruction and completion.}
\label{figure:figure_recon_comp}
\end{figure*}

In Figure~\ref{figure:figure_recon_comp}, we visualize the results of image reconstruction (a,b) and image completion (c-h).
As we can see, for face images, our proposed CVAE models are in general good at reconstructing and predicting the occluded region in unseen images (from testing set).
However, for bird images, vanilla CVAE model had significant failures in general. This agreed with the previous results in attribute-conditioned image generation.

In addition, to demonstrate the significance of attribute-conditioned modeling, we compared our vanilla CVAE and disCVAE with unconditional VAE (attribute is not given) for image reconstruction and completion.
It can be seen in Fig.~\ref{figure:figure_recon_comp}(c)(d), the generated images using attributes actually perform better in terms of expression and eyewear (``smiling'' and ``sunglasses'').

For quantitative comparisons, we measured the pixel-level mean squared error on the entire image and occluded region for reconstruction and completion, respectively. 
We summarized the results in Table~\ref{tab:table_recon_comp} (mean squared error and standard error).
The quantitative analysis highlighted the benefits of attribute-conditioned modeling and the importance of layered modeling.
%

\begin{table}[t]
\small
\centering
\caption{
Quantitative comparisons on face reconstruction and completion tasks. 
}
\begin{tabular}{c|c|c|c|c|c|c}
Face &  Recon: full & Recon: fg  & Comp: eye  & Comp: mouth & Comp: face & Comp: half \\
\hline
VAE & 11.8 $\pm$ 0.1 & 9.4 $\pm$ 0.1 & 13.0 $\pm$ 0.1 & 12.1 $\pm$ 0.1 & 13.1 $\pm$ 0.1 & 21.3 $\pm$ 0.2\\
CVAE & 11.8 $\pm$ 0.1 & 9.3 $\pm$ 0.1 & 12.0 $\pm$ 0.1 & 12.0 $\pm$ 0.1 & 12.3 $\pm$ 0.1 & 20.3 $\pm$ 0.2\\
disCVAE & 10.0 $\pm$ 0.1 & 7.9 $\pm$ 0.1 & 10.3 $\pm$ 0.1 & 10.3 $\pm$ 0.1 & 10.9 $\pm$ 0.1 & 18.8 $\pm$ 0.2\\
\hline\hline
Bird & Recon: full & Recon: fg & Comp: $8\times 8$ & Comp: $16\times 16$ & &\\
\hline
VAE & 14.5 $\pm$ 0.1 & 11.7 $\pm$ 0.1 & 1.8 $\pm$ 0.1 & 4.6 $\pm$ 0.1 & &\\
CVAE & 14.3 $\pm$ 0.1 & 11.5 $\pm$ 0.1 & 1.8 $\pm$ 0.1 & 4.4 $\pm$ 0.1 & &\\
disCVAE & 12.9 $\pm$ 0.1 & 10.2 $\pm$ 0.1 & 1.8 $\pm$ 0.1 & 4.4 $\pm$ 0.1 & &\\
\end{tabular}
\label{tab:table_recon_comp}
\end{table}


\section{Conclusion}
To conclude, this paper studied a novel problem of attribute-conditioned image generation and proposed a solution with CVAEs.
Considering the compositional structure of images, we proposed a novel disentangling CVAE (disCVAE) with a layered representation. 
Results on faces and birds demonstrate that our models can generate realistic samples with diverse appearance and especially disCVAE significantly improved the generation quality on bird images.
To evaluate the learned generation models on the novel images, we also developed an optimization-based approach to posterior inference and applied it to the tasks of image reconstruction and completion with quantitative evaluation.\\

\textbf{Acknowledgement.}
This work was supported in part by NSF CAREER IIS-1453651, ONR N00014-13-1-0762, Sloan Research Fellowship, and a gift from Adobe. We acknowledge NVIDIA for the donation of GPUs. We also thank Yuting Zhang, Scott Reed, Junhyuk Oh, Ruben Villegas, Seunghoon Hong, Wenling Shang, Ye Liu, Kibok Lee, Lajanugen Logeswaran, Rui Zhang, Changhan Wang and Yi Zhang for helpful comments and discussions.

\clearpage

\appendix

\section{Derivation of disCVAE objective}
\label{sec-app-deriv}
We provide a detailed derivation of the objective function for disentangling CVAE (disCVAE). Similarly to the vanilla CVAE, we have $x$ and $x_F$ as input image (full, foreground), $g$ as foreground mask, $y$ as attribute labels, and $z = [z_F, z_B]$ as latent variables ($z_F$ for foreground and $z_B$ for background). 

The joint conditional log-likelihood of $x$, $x_F$ and $g$ given $y$ can be written as follows:
\begin{align}
&\log p_{\theta} (x_F, g, x|y)\\
= &\;\mathbb{E}_{q_\phi (z_F, z_B|x_F, g, x, y)} \big[\log p_\theta (x_F, g, x|y)\big]\nonumber\\
= &\;\mathbb{E}_{q_\phi (z_F, z_B|x_F, g, x, y)} \big[ \log p_\theta (x_F, g, x, z_F, z_B|y) - \log p_\theta (z_F, z_B|x_F,g,x,y) \big]\nonumber\\
= &\; \textit{KL} ( q_\phi (z_F, z_B|x_F,g,x,y) || p_\theta (z_F, z_B|x_F,g,x,y) )\nonumber\\
&+ \underbrace{\mathbb{E}_{q_\phi (z_F, z_B|x_F,g,x,y)} \big[ \log p_\theta (x_F,g,x,z_F, z_B|y) - \log q_{\phi} (z_F, z_B|x_F,g,x,y) \big]}_{\triangleq \mathcal{L}_{\text{disCVAE}} (x_F,g,x,y;\theta, \phi)},\nonumber
\end{align}

Based on the disentangling assumptions, we write the generation model by
\begin{align}
p_\theta (x_F, g,x, z_F, z_B|y) =&\; p_\theta (x|z_F, z_B, y) p_\theta (x_F,g|z_F, y) p_\theta (z_F) p_\theta (z_B),
\end{align}
the recognition model by
\begin{align}
q_\phi (z_F, z_B|x_F,g,x,y) =&\; q_\phi (z_B|z_F, x_F, g, x, y) q_\phi (z_F|x_F, g, y)
\end{align}
and thus the variational lower bound $\mathcal{L}_{\text{disCVAE}} (x_F,g,x,y;\theta, \phi)$ is given by
\begin{align}
&\; \mathcal{L}_{\text{disCVAE}} (x_F,g,x,y;\theta, \phi)\nonumber\\
=&\; -\textit{KL} ( q_\phi (z_F|x_F,g,y) || p_\theta(z_F) )
- \mathbb{E}_{q_\phi (z_F|x_F, g,y)} \big[ \textit{KL} (q_\phi (z_B|z_F, x_F, g, x, y) || p_\theta (z_B) ) \big]\nonumber\\
&\; + \mathbb{E}_{q_\phi (z_F|x_F, g, y)} \big[ \log p_\theta (x_F,g|z_F, y) \big] + \mathbb{E}_{q_\phi(z_F, z_B|x_F, g, x, y)} \big[ \log p_\theta (x|z_F, z_B, y) \big]\nonumber\\
=&\; -\textit{KL} ( q_\phi (z_F|x_F,g,y) || p_\theta(z_F) ) 
- \mathbb{E}_{q_\phi (z_F|x_F,g, y)} \big[ \textit{KL} ( q_\phi (z_B|z_F, x_F, g, x, y) || p_\theta (z_B) ) \big]\nonumber\\
&\ - \mathbb{E}_{q_\phi (z_F|x_F, g, y)} \big[ L(\mu_{\theta_F}(y, z_F), x_F) + \lambda_g L(s_{\theta_g}(y, z_f), g) \big]\nonumber\\
&\ - \mathbb{E}_{q_\phi(z_F, z_B|x_F, g, x, y)} L(\mu_{\theta} (y, z_F, z_B), x)\nonumber\\
\end{align}

In the last step, we assumed that $\log p_\theta(x_F, g|z_F, y) = \log p_\theta(x_F|z_F, y) + \lambda_g \log p_\theta(g|z_F, y)$, where $\lambda_g$ is a hyperparameter when decomposing the probablity $p_\theta (x_F,g|z_F, y)$.
Here, the third and fourth terms are rewritten as expectations involving reconstruction loss (e.g., $\ell_2$ loss) or cross entropy.

\section{Network Architecture for disCVAE}
\label{sec-app-netarch}
\begin{figure*}[t]
\includegraphics[width=\linewidth]{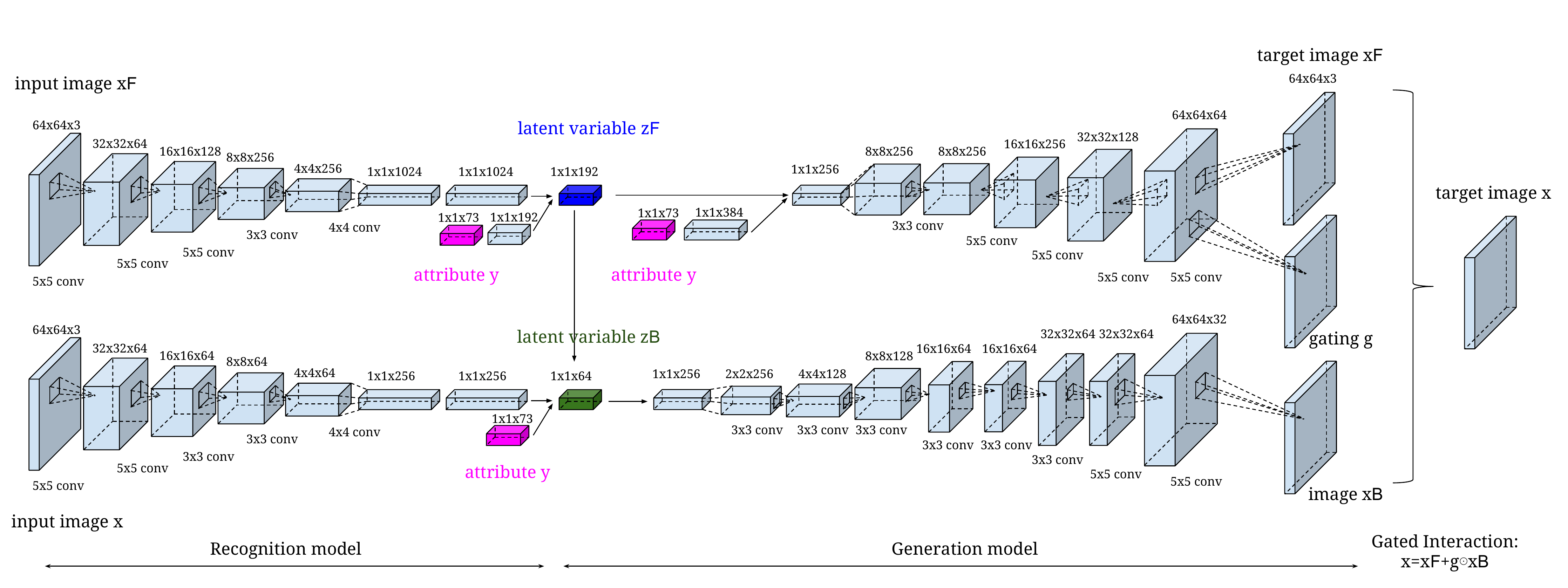}
\caption{
Network Architecture for disentangling CVAE
}
\label{figure:figure_pipeline}
\end{figure*}

As we visualize in Figure~\ref{figure:figure_pipeline}, disCVAE consists of four convolutional neural networks (one for foreground and the other for background for both recognition and generation networks).

The foreground encoder network consists of 5 convolution layers, followed by 2 fully-connected layers (convolution layers have 64, 128, 256, 256 and 1024 channels with filter size of $5 \times 5$, $5 \times 5$, $3 \times 3$, $3 \times 3$ and $4 \times 4$, respectively; the two fully-connected layers have 1024 and 192 neurons). The attribute stream is merged with image stream at the end of the recognition network.
The foreground decoder network consists of 2 fully-connected layers, followed by 5 convolution layers with 2-by-2 upsampling (fully-connected layers have 256 and $8 \times 8 \times 256$ neurons; the convolution layers have 256, 256, 128, 64 and 3 channels with filter size of $3 \times 3$, $5 \times 5$, $5 \times 5$, $5 \times 5$ and $5 \times 5$. The foreground prediction stream and gating prediction stream are separated at the last convolution layer.

We adopt the same encoder/decoder architecture for background networks but with fewer number of channels.  For better modeling on the background latent variable $z_B$, we introduce attribute $y$ and foreground latent variable $z_F$ into the background encoder network, which also agrees with the assumption made in the derivation ($q_\phi (z_B|z_F, x_F, g, x, y)$). Here, the connection from foreground latent variable $z_F$ to background latent variable $z_B$ only exists in the recognition model.

Note that encoder networks are only used during the training stage. Once trained, we can generate images using decoder networks only.

\section{Quantitative Analysis: Attribute Similarity, Labeled Faces in the Wild}
\label{sec-app-attrpred}
In order to measure the performance quantitatively in the attribute space, we propose to evaluate whether the generated samples exactly capture the condition (attributes).
Therefore, we trained a separate convolutional neural network from scratch as attribute regressor using image-attribute pairs in the training set. The attribute regressor shares almost the same architecture with the auxiliary recognition model used in generative training. As a reference, the attribute regressor achieves 14.51 mean squared error (MSE) and 0.98 cosine similarity on the test set.

For each attribute vector in the test set (reference attribute), we randomly generate 10 samples and feed them into the attribute regressor. We then compute the cosine similarity and mean squared error between the reference attribute and the predicted attributes from generated samples. As a baseline method, we compute the cosine similarity between the reference attribute and the predicted attributes from nearest neighbor samples (NN).
Furthermore, to verify that our proposed method does not take unfair advantage in evaluation due to its somewhat blurred image generation, we add another baseline method (NN$_{blur}$) by blurring the images by a 5 x 5 average filter.

As we can see in Table~\ref{tab:tab_attrpred}, the generated samples are quantitatively closer to reference attribute in the testing set than nearest attributes in the training set.
In addition, explicit foreground-background modeling produces more accurate samples in the attribute space.

\begin{table}[thbp]
\small
\centering
\caption{
Quantitative comparisons on attribute-conditional image generation.
The best out of 10 samples are evaluated by the cosine similarity and mean squared error in the attribute space.
We use a pre-trained convolutional neural network to predict attributes of generated samples.
}
\begin{tabular}{c|c|c}
Model & Cosine Similarity & Mean Squared Error \\
\hline\hline
NN & 0.8719 & 21.88 \\
NN$_{blur}$ & 0.8291 & 28.24\\
CVAE &  0.9054 & 17.20 \\
disCVAE & 0.9057 & 16.71 \\
\end{tabular}
\label{tab:tab_attrpred}
\end{table}

\clearpage

\bibliographystyle{splncs03}
\bibliography{references}
\end{document}